\documentclass{article}

\usepackage{microtype}
\usepackage{graphicx}
\usepackage{subfigure}
\usepackage{booktabs} 

\usepackage{hyperref}

\usepackage[accepted]{icml2021}

\icmltitlerunning{Enhancing Reinforcement Learning with discrete interfaces}

\begin{document}

\twocolumn[
\icmltitle{Enhancing Reinforcement Learning with discrete interfaces\\to learn the Dyck Language}



\icmlsetsymbol{equal}{*}

\begin{icmlauthorlist}
\icmlauthor{Florian Dietz}{lsv}
\icmlauthor{Dietrich Klakow}{lsv}
\end{icmlauthorlist}

\icmlaffiliation{lsv}{Spoken Language Systems (LSV), Saarland University, Germany}

\icmlcorrespondingauthor{Florian Dietz}{fdietz@lsv.uni-saarland.de}
\icmlcorrespondingauthor{Dietrich Klakow}{dietrich.klakow@lsv.uni-saarland.de}

\icmlkeywords{Machine Learning, Artificial Intelligence, Dyck, Reinforcement Learning, ICML}

\vskip 0.3in
]



\printAffiliationsAndNotice{}  

\begin{abstract}

Even though most interfaces in the real world are discrete, no efficient way exists to train neural networks to make use of them, yet. We enhance an Interaction Network (a Reinforcement Learning architecture) with discrete interfaces and train it on the generalized Dyck language. This task requires an understanding of hierarchical structures to solve, and has long proven difficult for neural networks. We provide the first solution based on learning to use discrete data structures. We encountered unexpected anomalous behavior during training, and utilized pre-training based on execution traces to overcome them. The resulting model is very small and fast, and generalizes to sequences that are an entire order of magnitude longer than the training data.

\end{abstract}

\section{Introduction}

Neural Networks can be made more powerful by giving them access to external interfaces for storing and accessing data. Fully differentiable MANN systems, such as the Neural Turing Machine introduced by \citet{graves2014neural}, can generally be trained much more efficiently than those with discrete data structures. If an effective way to integrate discrete interfaces into a neural network could be found, it would have numerous advantages:

\begin{itemize}
	\item \textbf{Modelling Power} Many types of interfaces are inherently discrete and can not be replaced with differentiable alternatives. This includes API calls and user-interactions. With a discrete interface, a neural network could be given access to databases like Wikipedia, to look up arbitrary information as needed. Access to common-sense information like this is likely to increase the predictive power of the model.
	\item \textbf{Speed} Discrete control structure are much faster to execute than differentiable structures, and the cost is often independent of the size\citep{zaremba2015reinforcement}. In contrast, differentiable memory structures get slower to train and execute the larger they are.\footnote{Some architectures avoid this problem by representing the data in different ways, but they usually suffer other drawbacks as a result. For example, Neural Random Access Machines\citep{kurach2015neural} can change the size of their memory without retraining the model, but their training and execution speed are dependent on the memory size. They can use discretization during inference to avoid this, but only at the cost of performance).}
	\item \textbf{Robustness} Recurrent neural networks often fail to generalize to longer sequence lengths than those that were encountered during training, because small errors accumulate over time. Discrete control structures do not have this problem.
	\item \textbf{Explainability} Discrete data is much easier to understand than continuous data. Making AI algorithms more explainable to human users has become very important for industrial applications in recent years.
	\item \textbf{Modular Integration} Training networks with differentiable memories has been noted to be very difficult\citep{suzgun2019memory,kurach2015neural}. This precludes the use of these differentiable data structures as components in larger models, because it would make the training process too unreliable and expensive.\footnote{It is possible that this problem disappears as the technology matures and more effective ways to tune these networks are found, but for now it requires many trials with different hyperparameters and random seeds in order to find one that works.}
\end{itemize}

Because discrete interfaces are not differentiable, another way to train them needs to be found. The obvious candidate for this is Reinforcement Learning. \citet{zaremba2015reinforcement} have attempted to integrate discrete control structures into a neural network by using a Reinforcement Learner that learns to operate on memory tapes. Their model succeeds on a wide range of simple algorithmic tasks, but fails to learn effective memory access patterns for more complex problems.

In this paper, we explore this topic further: We tackle a more difficult problem, the \textbf{generalized Dyck Language}, using a more generic architecture. We chose the generalized Dyck Language for our task because it expresses the core of context free languages\citep{suzgun2019memory}, which require understanding hierarchical data. Humans naturally think in terms of hierarchies in many situations, but existing neural networks are unable to effectively learn tasks that require understanding hierarchical data\citep{chomsky2002syntactic}.

For our model, we use an \textbf{Interaction Network}\citep{dietz2020interaction}, which offers some additional abilities over a reinforcement learning neural network. An Interaction Networks (IN) is a type of cognitive architecture based on combining multiple independent neural networks with discrete control structures and independent memory Nodes. It uses Reinforcement Learning to train separate smaller neural networks, and can learn to combine them to solve more complex tasks. Concretely, our Interaction Network contains two semi-independent neural networks: A reinforcement learner that decides which actions to perform, and a feedforward network that learns a mapping between symbols.

Our experiments show that Reinforcement Learning has difficulties learning Dyck and results in unusual error modes. However, trials differ substantially based on the random initialization of the model. We therefore tested pre-training the Reinforcement Learner on small samples of execution traces. This leads to good results that generalize well to much higher sequence lengths.

\section{Related Work}

\subsection{Memory Augmented Neural Networks (MANN)}

There are several types of interfaces that can be made available to an neural network, but the most basic and important of them is giving a neural network an external memory. These systems are called Memory Augmented Neural Networks (MANN), and they improve upon more traditional memory mechanisms such as LSTMs. Examples are \citet{weston2014memory}, \citet{graves2014neural}, \citet{grefenstette2015learning}, \citet{sukhbaatar2015end}, \citet{graves2016hybrid}, \citet{yang2016lie}, and \citet{kaiser2015neural}.

\subsection{Program Induction and Program Synthesis}

Our aim is to train a system to solve the Dyck task with explicit rules. There are two ways to do this: Program Induction or Program Synthesis. In Program Induction, a neural network is trained to simulate the behavior of a program directly. In Program Synthesis, a neural network is trained to output a formal description of a program, which is then executed to solve the problem. Program Induction is the default approach, as all neural networks that try to solve programming-related tasks without explicitly creating a program are implicitly performing Program Induction. Examples for Program Synthesis are \citet{parisotto2016neuro} and \citet{nye2020learning}.

Our approach is a mix of both. On the one hand, the Interaction Network performs a sequence of discrete actions that can be viewed as the elementary statements of a programming language. On the other hand, this sequence of actions is generated one step after the other, at runtime.

\subsection{Reinforcement Learning and discrete control structures}

\citet{zaremba2015reinforcement} examine the use of Reinforcement Learners in conjunction with discrete interfaces to solve simple algorithmic tasks. They used memory tapes as the discrete interface, which effectively gave their system the capabilities of a Turing Machine.

The resulting model successfully learns to solve problems such as Copy, RepeatCopy and Reverse. They found that their model either converged in under 20.000 updates or got stuck in a local optimum and failed to converge. Our own findings were similar: The solution is either found quickly, or not at all.

\subsection{Interaction Networks}

An \textbf{Interaction Network (IN)} is a meta-learning system that can train multiple independent neural networks and combine them as necessary to solve a problem.

The architecture of the Interaction Network that is used in this paper is very simple and uses only a subset of the abilities of an Interaction Network. We will therefore only outline the key components, and refer to the original paper \citet{dietz2020interaction} for details. The specific architecture of the IN used in our experiments is described in the Methods section.

Structurally, an IN consists of the following:

\begin{itemize}
	\item \textbf{Nodes}, which hold a fixed-size data tensor persistently until an action explicitly overwrites it.
	\item \textbf{Processing Units (PU)}, which are neural networks that map from input Nodes to output Nodes.
	\item A \textbf{Control Unit (CU)}, which is a Reinforcement Learner.
	\item Control structures, such as interfaces to tasks and memory storage mechanisms.
\end{itemize}

The IN does not have a fixed input-output mapping. Instead, it is a lifelong-learner that runs in an infinite loop. On each iteration, the CU chooses an action to perform. Possible actions include obtaining a new input from the task interface, submitting an output, moving data from one Node to another, or executing a Processing Unit.

When a Processing Unit is executed, it uses whatever values are currently stored in its input Nodes as its inputs and overwrites its output Nodes with the results. Because of this, and because the order of PU execution is determined by the CU and not known in advance, the computational graph is constructed dynamically. This makes the closed-form expression of the IN as a whole very hard to read, as it is a recursive structure full of conditional statements. Each individual CU and PU is based on simple feedforward networks, but the way in which they interact leads to a complex formula.

The IN architecture used for the experiments in this paper uses only a single PU. Nevertheless it is not easy to give a closed-form expression because the CU is separate from the PU and its actions decide which data the PU receives as input.

External interfaces can be attached to an IN. These interact with Nodes to receive and deliver data to the rest of the IN. The CU decides when and how to use external interfaces. The CU may use external interfaces and PU's in any order, and may even decide to use them many times in a row, or not at all. The reward of the Reinforcement Learner is used to train the CU to use PU's and external interfaces in a sensible order.

Interaction Networks have a large number of interacting components, which makes them difficult to implement. See \citet{dietz2020interaction} for code and implementation details.\footnote{\citet{zaremba2015reinforcement} noted similar difficulties in combining external interfaces with a Reinforcement Learner. Neural networks without external interfaces are much easier to write nowadays than they were a decade ago, because of libraries like PyTorch and TensorFlow. No libraries for combining external interfaces with neural networks exist, yet.}

\subsection{The Dyck language}

The generalized Dyck language $D_n$ consists of strings with balanced pairs of brackets, such as "[]()\{()[]\}". The $n$ in $D_n$ specifies the number of different types of brackets that may appear in the string.

For $n=1$, the Dyck language can be modelled merely by counting the number of brackets that still need to be closed. For $N>1$, it becomes necessary to keep track of the sequence of brackets, which is much more difficult.

Dyck words can be generated recursively. There are different ways to parameterize this, and different papers use different methods. Additionally, papers differ in the type of task: Some require the model only to find the next valid closing bracket of a Dyck prefix, some require a valid completion of the word, and some require the shortest possible completion.

As a result, different papers use slightly different procedures\citep{deleu2016learning, sennhauser2018evaluating, skachkova2018closing, yu2019learning, hahn2020theoretical}. However all of these variants still capture the core difficulty of the task: The algorithm needs to keep track of  the sequence of  brackets.

In this paper, we are only interested in finding a solution to Dyck tasks that generalizes perfectly to much longer words. For this reason, these minor differences are unimportant. A solution to one variant is as powerful as a solution for any other variant.

We generate Dyck words dynamically by first selecting a requisite length for the input sequence and then generating a valid Dyck prefix of that length. On each step, the probability of closing the last open bracket (if one exists) is 0.5, and otherwise a random new bracket is opened. Once the required length has been reached, the remaining Dyck brackets that need to be closed are used as the target sequence of the model. Both the input and the output sequence are terminated by a special termination symbol $\epsilon$.

\subsection{Existing solutions to Dyck}

Solutions to $D_1$ can be found using simple counting and are therefore not interesting for testing if a model understands hierarchical data. The more interesting task is to complete a Dyck word for $D_{>1}$. We are interested in testing if the solution generalizes to longer words after training.

It is possible to train simple RNNs to complete Dyck words of fixed length. However these solutions fail to generalize to longer word lengths. This suggests that the algorithm only learns statistical correlations and does not capture the true logic behind generating the Dyck words. \citet{suzgun2019memory} provides the first demonstration of neural networks capable of solving the generalized Dyck language. Their solution is based on memory-augmented Recurrent Neural Networks (MARNNs).

MARNNs use an integrated fully differentiable structure that simulates the behavior of a stack. On each iteration, the MARNNs perform both push and pop operations on the stack simultaneously, each with a learned weight, and store a weighted superposition of both results on the stack. This makes the decision whether to push or pop learnable by backpropagation. Their models were trained on sequence lengths up to 50, and were tested on sequence lengths up to 100. Many of them achieved optimal performance on the training set and maintained this even on the test set. This shows that they generalized correctly and learned to use their internal stack to learn the logic behind the Dyck language.

To our knowledge, our own experiments are the first solution based on a neural network with discrete interfaces.

\section{Methods}

We ran two groups of experiments. Both used the same task definition and IN architecture. In the first, the CU was initialized randomly. In the second, the CU was pre-trained to be close to the correct solution using execution traces, then randomly modified with noise.

It is important to note that all of these experiments ran on a laptop. The neural networks trained here are far smaller than most used in contemporary literature and had only a few thousand parameters each. This was a deliberate choice, because understanding the logic behind Dyck tasks should not require too many parameters. More importantly, a smaller model is more useful than a big one, since it is both faster to train and faster to run.

\subsection{Task}

An Interaction Network is an online learning system. Instead of using an explicit training set and test set, the IN is trained on a dynamically created curriculum of tasks of increasing complexity. Learning in such a manner is much closer to the way humans learn, and has been shown to speed up learning and improve robustness\citep{bengio2009curriculum}. The performance on each level of the curriculum is tracked over time. More difficult tasks are unlocked once less difficult ones reach a 95\% success rate. The algorithm will sometimes revisit lower difficulties at random, to verify that they still work.

This makes it possible to track after how many iterations a given level of task difficulty can be solved reliably. It also enables us to check whether this performance remains stable or can experience catastrophic forgetting when the network trains on higher difficulties. Since the number of possible Dyck words grows exponentially with the length of the word, the probability of training on duplicates is very low. Therefore, there is no risk of overfitting.


The curriculum at first uses only a small increase in difficulty in order to train the model gradually, but later it introduces large jumps in difficulty in order to test the model's ability to generalize: After solving tasks with input sequence length 10, the next value in the curriculum uses an input length of 100, and the one after that uses a length of 1000. To our knowledge, no previous work has attempted testing networks with such high spikes in difficulty. Usually the test set has at most twice as high a sequence length as the training set, not an entire order of magnitude in difference.

This large jump in difficulty is deliberate. It is possible to train a neural network to achieve good performance on Dyck completion tasks where the tested sequence length is only a little bit higher than during training\citep{yu2019learning}. Presenting the network with a large jump in difficulty means that it can only succeed if it learned the real underlying algorithm behind Dyck, and not statistical dependencies.

\subsection{Model}

\begin{figure}
	\centering
	\fbox{
		\includegraphics[width=0.5\columnwidth,height=100pt]{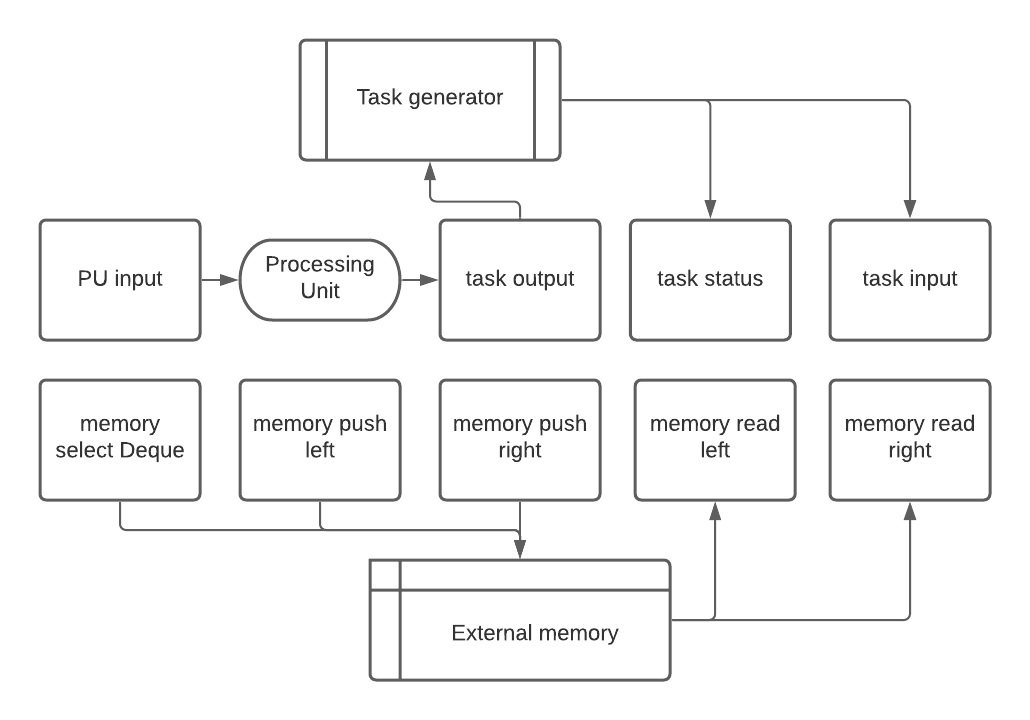}
	}
	\caption[list=off]{This graphic shows the Nodes, interfaces and Processing Unit used by the IN, and their connections to the task generator and the discrete external memory. The nine central rectangles are Nodes. The top and bottom rectangle are external interfaces. The oval is the Processing Unit.}
	\label{fig:in_architecture}
\end{figure}

Figure \ref{fig:in_architecture} shows an abstracted view of the Interaction Network used in these experiments. The IN uses nine Nodes to store data. Each of these Nodes can hold one tensor, and that tensor persists until the CU takes an action to overwrite it. The task uses two Nodes to deliver input to the IN. One indicates when a new task is started, the other delivers the currently selected input element. The CU can use one action to tell the Task generator to show the next input element, and another action for the previous input element. The different symbols are encoded as random hashes onto a fixed-sized vector. This encoding is harder to learn than a one-hot encoding, but it makes the system more flexible because new symbols may be introduced later, at any time.

The IN has one Processing Unit, which is a feedforward network. In order to solve the task, this network has to learn the mapping between opening and closing brackets. Whenever the value of the "PU input" Node changes, the PU executes with this input and writes its output into the "task output" Node. This submits the output symbol.

Whenever an output is submitted, the Task generator applies an MSE loss to the submitted value (which trains the PU) and a reward to the IN (which trains the CU by reinforcement learning). This means that the IN has two independent trainable components: The CU and the PU. The CU decides how to move data between Nodes and how to request input elements and interact with the memory. It is a Reinforcement learner. The PU is a simple feedforward network that learns a mapping between symbols. It learns using the MSE loss.

This division of concerns between what to do (CU) and how to do it (PU) is a design decision of Interaction Networks. It is not strictly necessary for this problem, since \citet{zaremba2015reinforcement} has shown that RL systems can also learn these sort of tasks with a single neural network. However, the ability to split a problem into independently learnable sub-problems is an important aspect of human intelligence, which is why Interaction Networks are designed in this modular fashion.

The external memory used in this Interaction Network is a combination of hash table and DeQue. This is a general-purpose data structure that can be used for many purposes. It is more complex than it would need to be for solving the Dyck language, since a stack is known to suffice. This is a deliberate design decision to check if the RL algorithm gets distracted if it has more actions available than it needs.

The external memory contains a collection of DeQues and uses a hash value to switch between them. Selecting a previously unused hash switches to an empty DeQue and effectively resets the memory. One Node is used for selecting the hash, and one each for appending a tensor to the left or right of the selected Deque. Two Nodes are used to make the leftmost and rightmost element of the currently selected DeQue available to the network.

\begin{figure}
	\centering
	\fbox{
		\includegraphics[width=0.5\columnwidth,height=100pt]{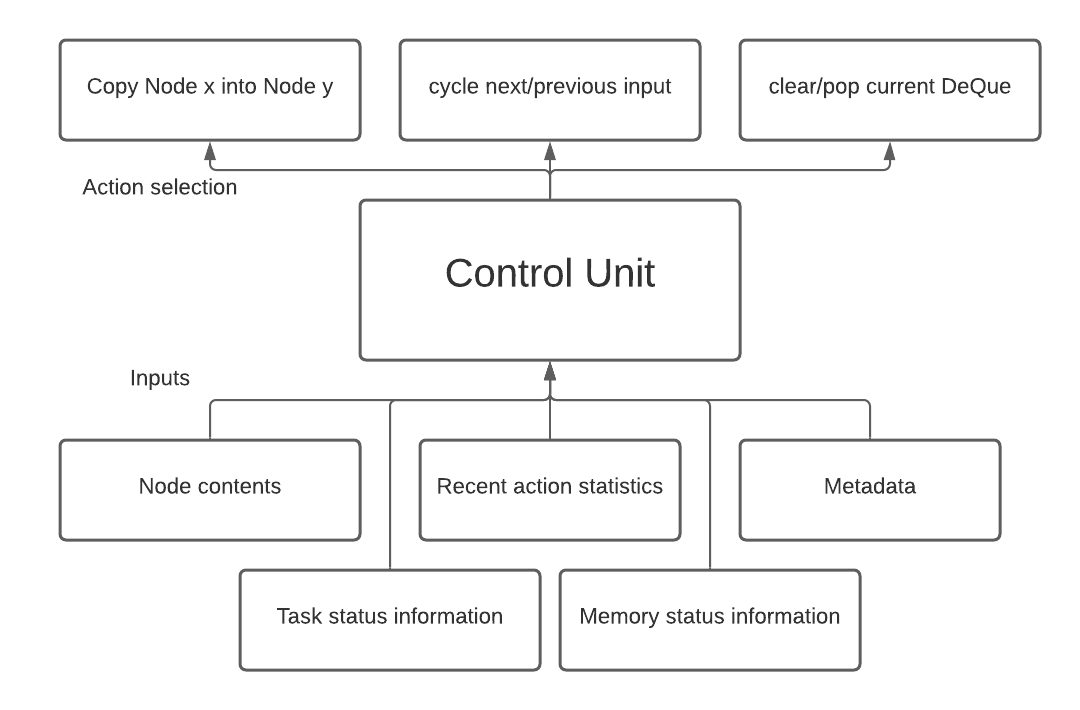}
	}
	\caption[list=off]{This graphic shows the different inputs the CU receives, and the actions it can choose between.}
	\label{fig:cu_io}
\end{figure}

The CU takes the current state of the IN as input, and chooses which action to take next. This is shown in \ref{fig:cu_io}. The CU's inputs include the current states of the Nodes, as well as information about the action taken on the last iteration, indicators of the Task generator (length and current position of the input sequence) and the external memory (the size of the DeQue), as well as some metadata about the network.

The CU uses the REINFORCE algorithm to select the next action to execute. Available actions are interactions with the Task generator or the external memory, as well as operations that copy a tensor from one Node into another. Copying can occur between some but not all pairs of Nodes. We made some copying operations available as actions that we know are not necessary for solving Dyck, to check if these distract the IN.\footnote{Making only the actions available that we know are required for a perfect solution would defeat the point of the experiment. It would shrink the search space of the CU so much that even an exhaustive search would be practically guaranteed to converge.}

For illustrative purposes, here is a comparison to the architecture used by \citet{zaremba2015reinforcement}:

\begin{itemize}
	\item The CU corresponds to the Direct Access Controller. It does not use an LSTM, because this can cause the controller to get stuck in a local optimum as it comes to rely too much on the LSTM and neglects to learn how to use the external memory properly.
	\item The CU takes many inputs: The values of every Node, as well as various additional metadata.
	\item The Direct Access Controller interacts with a memory tape, while the IN uses a hybrid data structure combining a hash and a DeQue. We chose this data structure because programmers use Hashtables and DeQues a lot in practice, while Turing Machines are mostly just used for their historical value and see little use in practical applications. The bybrid data structure can represent arbitrary data structures by using an element from the active DeQue as the key of another DeQue, so that it effectively acts as a pointer.
	\item The CU can perform movement actions to copy a tensor from one Node into another. Something similar was also used in \citet{zaremba2015reinforcement}. They at first used an LSTM which had to learn the mapping between symbols and could not copy symbols from one location to another. They then introduced the Direct Access Controller, which could copy symbols and achieved better performance.
\end{itemize}

\textbf{Reward tuning} Reward tuning is an important part of Reinforcement Learning. We tried two different models:

\begin{itemize}
	\item Standard temporal-difference reward: A reward of either 0 or 1 is given at the end of a task and is applied to the last step of the task. On all other iterations, the CU is trained via temporal-difference learning. We tried several values for the reward discount factor.
	\item Whole-episode rewards: No temporal-difference learning is used. Instead, all actions taken during a task receive the task reward directly.
\end{itemize}

\textbf{Random exploration} Another important aspect of tuning Reinforcement Learners is random exploration. We configured the CU to use different types of random behavior on different trials. In some tasks it would act completely deterministically until the task was done. In others it would have a fixed probability of taking a random action on each iteration. Since longer Dyck tasks take more actions to solve, we expected the effects of random actions to be more pronounced on longer tasks.

\subsection{Pre-training the model}

Learning the Dyck completion task with randomly initialized models did not yield solutions that generalize to longer sequence lengths. See the Results section for details. One finding that jumped out was that the network often behaved wildly differently based on changes in its initialization and the first examples it encountered.

A good initialization seems crucial for convergence. If the model starts off with a bias towards an ineffective strategy, it is more likely to get stuck. Most random initializations seem to cause the network to converge to simple heuristics that work well on short sequences, but do not generalize well.

In order to test if a sensibly preconfigured network would be more likely to converge to the optimal strategy, we used to following approach: First, we pre-train the network on execution traces to follow a manually defined optimal strategy that solves the Dyck task. We train on this task until the network achieves good performance on short sequence lengths. At this point, the pre-trained model does not generalize to longer sequences yet, because the training is stopped early and it is not fully converged yet. Then we introduce additional noise into the pre-trained network to reduce its ability further.

The pre-trained network generated in this way does not manage to solve even short sequences reliably. However, the pre-training biases the CU weights in such a way that it is more likely to find the optimal strategy during training, and less likely to get stuck in a local optimum.

\citet{reed2015neural} has surface similarities to our approach. Their Neural Programmer-Interpreters is a neural network that can compose smaller components to solve complex problems. They used execution traces as their learning method. Their results demonstrate that a model that is trained entirely on execution traces can be very powerful and general. Unfortunately, execution traces can be hard to generate for some types of tasks, which limits the usefulness of a system that relies on them too much. That is why we take a mixed approach: We only use a small number of simple execution traces to pre-train a model, then rely on normal training data to fine-tune it.

\section{Results}

\subsection{Why the numbers don't matter}

We do not report numerical values from our experimental results, because we have found these to be highly misleading. We have manually investigated the action sequences taken by the network in our experiments,\footnote{These experiments were performed using the code for Interaction Networks in \citet{dietz2020interaction}, which include a GUI that makes it easy to inspect the behavior of the network both statistically and on a step-by-step basis.} and found that there was little correlation between the usual KPIs and a human evaluation of how well the algorithm learned Dyck.

Our main goal was to train a system that understands the underlying logic behind the Dyck task. This problem is binary: The algorithm has either understood the logic, or it has not. Since we know what a perfect solution to Dyck should look like, we can manually analyze the execution traces of the trained system to find out how close it comes to the true solution.\footnote{There are of course several possible true solutions, and in theory we risk missing a potential solution that the AI can learn that never occurred to any human. However we find this to be very unlikely in practice, since the execution traces frequently contain actions that are quite obviously useless or counterproductive.} Our manual evaluation of the learned algorithm rarely matches the numerical quality of the result. This is because the network learns spurious correlations that do not generalize well.

Because of this, there is little point in reporting the exact numerical performance of each run. Instead, we describe what the different reasons for failure were based on manual analysis of the results, and what the impact of different parameters was.

\subsection{With randomly initialized models}

When using randomly initialized models, without pre-training on execution traces, performance was generally poor. The reason for the poor performance was always that the network got stuck in a local optimum. However, the reasons for that local optimum differed between trials. Just changing the random seed can alter the results.

\subsubsection{Failure Modes}

\textbf{Incorrectly trained Processing Unit} One problem was that the Processing Unit was trained wrong. The actions of the CU determine the training data of the PU, while the performance of the PU determines how well the CU can learn how to use the PU. This cyclic dependency can cause difficulty during training, as noted in the introductory paper for Interaction Networks\citep{dietz2020interaction}.

We fixed this by adding a new task as the first part of the curriculum, which was purely about matching opening brackets to closing brackets. This allowed the CU to train the PU. After this, the CU is able to use the trained PU to learn the Dyck task more efficiently, since it can now use the PU as a subroutine for matching brackets and does not have to learn this itself.\footnote{This subtask was not necessary in the experiments with a pre-trained model.}

\textbf{Misleading tasks in the curriculum} Starting the curriculum with a too low difficulty actually harmed the learning process. Very short Dyck words can be completed without using a stack at all. As a result, the CU learned a strategy that solved short sequences perfectly, but performed poorly on longer task sequences. Starting the curriculum at higher sequence lengths fixed this problem.

\textbf{Difficulty assigning blame for failure on higher sequence lengths} In cases where the IN did learn a correct solution for short sequences, the quality of the solution deteriorated as the task sequence length increased. The problem is that it is hard to “assign blame” for a failure. i.e. to determine which action caused the failure. For example, removing an item from the DeQue on the wrong end only leads to failure much later, when that item is supposed to be used. This problem becomes more pronounced the longer the task sequence is.

\subsubsection{Reward tuning}

Rewarding every action on the whole episode performed better than temporal-difference learning. Temporal-difference learning leads to steadily declining expected rewards, which gets worse the higher the sequence length is. Long sequences can cause the CU to deteriorate because there is only one fixed reward at the end of the episode, but each intermediate step drives down the expected reward of the action. The longer the sequence, the more the reward is diluted. This causes catastrophic forgetting: The CU mistakenly assumes that the actually correct action sequence is wrong and lowers its expected reward because the delay until the actual reward is too high. This causes the CU to attempt to use other actions. Since those other actions are actually bad, they break the entire chain of actions and cause the CU to fail to work on the longer sequence.

Temporal-difference learning also slows down the learning process. Unlike most types of neural network, an Interaction Network may decide on its own whether or not it wants to submit a result, and is not forced to do so on each iteration. This can lead to curious behavior. If the IN reaches a point where it knows it has made a mistake and is unlikely to succeed at the task, it starts performing random actions except those that would submit an output. This happens because random procrastinating actions at least still get a temporal-difference reward, while submitting a result is sure to result in a failure. This continues until the task times out. The network does eventually learn not to do this, but it takes a lot of time and a significant amount of training time can be wasted on this procrastinating behavior.

\subsubsection{Random exploration}

Random exploration had a mostly negative effect on training. It was unsurprising that random exploration is harmful on longer sequence lengths, since operating a stack properly is precise work and mistakes are difficult to undo. What was surprising was that random exploration only rarely helped on short sequence lengths, either. Performance was best when we configured the CU to be entirely deterministic and always take the most highly rated action.

We hypothesize that the reason for this is the highly regular nature of the task. Solving the Dyck task requires performing the exact same short sequence of actions repeatedly until the task is done. This regularity is so important that deviating from it can ruin performance even on short sequences.

\subsection{With Pre-training the model}

The pre-trained model learns more quickly and generalizes much better than the randomly initialized model. The model does still occasionally experience a fall into local optima when the curriculum introduces a higher sequence length. However, it always manages to recover and get back on track.

After the curriculum jumps from sequence length 10 to 100, our model performs adequately but not perfectly at first. From inspecting the results manually, it seems that the agent is mostly following the correct strategy, but occasionally makes mistakes due to slight errors that accumulate over longer sequences.

Crucially, unlike the model that was trained from scratch, the pre-trained model does not get stuck in local optima when it encounters these problems. It manages to correct itself, and eventually achieves full performance even on the tasks with sequence length 100. The same happens after increasing the difficulty from length 100 to length 1000.

Curiously, the agent sometimes over-corrects when learning these sequences. This causes it to lose performance for many consecutive task instances before it catches itself again. Further research is needed to investigate the cause of these over-corrections, and how to prevent it. We suspect that this was caused by the way reward signals are trained on. As mentioned, instead of using temporal-difference learning the CU receives identical reward signals for every iteration that was taken during a task. On longer task sequences, this causes a lot of similar entries to be added to the Experience Replay of the Interaction Network at the same time, which could bias the learning process.

The network did not achieve full convergence on the length 1000 task in the time we had allocated for the experiment, though it was on the way to get there. The jump from length 10 to 100, the convergence at length 100, and the jump from length 100 to length 1000 all demonstrate that the network generalizes well and learns the real logic underlying the Dyck task.

\section{Conclusion}

In conclusion, Reinforcement Learning combined with discrete interfaces can learn to understand the Dyck language, but only with pre-training on execution traces. Without pre-training, Reinforcement Learning is very susceptible to local optima.

If the RL agent can avoid falling into a local optimum, it can learn the optimal strategy and generalizes very well. It is also much smaller and faster than comparable models.

\section{Future Work}

Pre-training the model on execution traces proved crucial for finding a good model that generalizes well. For many tasks, it is possible to provide execution traces that solve at least simple task instances perfectly. Our experiments suggest that this is well worth the effort when it is possible to do so. However, there are other tasks where we do not know how to define sample execution traces manually, or where doing so is possible but expensive.


This begs the question if it is possible to replace pre-training on execution traces with a more practical alternative. Transfer learning in a multi-task environment could be a solution to this problem: Instead of training on Dyck directly, give a curriculum of tasks to the system that grows more complex on a semantic level, rather than just increasing sequence lengths. Start with simple tasks like just repeating one character, then teach it tasks that require lists, queues and stacks. A network that can perform other stack-processing tasks is a promising pre-trained candidate for learning Dyck.


\citep{fernando2017pathnet} did something similar with PathNet. It successfully demonstrated transfer learning between MNIST, CIFAR, SVHN supervised learning classification tasks, and a set of Atari and Labyrinth reinforcement learning tasks.

\bibliography{example_paper}
\bibliographystyle{icml2021}

\end{document}